\documentclass{article}
\usepackage{spconf,amsmath,graphicx}
\usepackage{subcaption}
\usepackage{todonotes}
\renewcommand{\paragraph}[1]{\noindent\textbf{#1}\quad}
\title{Multitask Learning for Low Resource Spoken Language Understanding}
\name{Quentin Meeus \textsuperscript{1,2}, Marie-Francine Moens \textsuperscript{1}, Hugo Van hamme \textsuperscript{2}\thanks{This research received funding from the Flemish Government under the “Onderzoeksprogramma Artificiele Intelligentie (AI) Vlaanderen” programme.}}
\address{
    \textsuperscript{1} KU Leuven, Dept. Computer Sciences CS-LIIR \\
    \textsuperscript{2} KU Leuven, Dept. Electrical Engineering ESAT-PSI\\
    quentin.meeus@esat.kuleuven.be}

\begin{document}
%\ninept
%
\maketitle
\begin{abstract}
We explore the benefits that multitask learning offer to speech processing as we train models on dual objectives with automatic speech recognition and intent classification or sentiment classification. Our models, although being of modest size, show improvements over models trained end-to-end on intent classification. We compare different settings to find the optimal disposition of each task module compared to one another. Finally, we study the performance of the models in low-resource scenario by training the models with as few as one example per class. We show that multitask learning in these scenarios compete with a baseline model trained on text features and performs considerably better than a pipeline model. On sentiment classification, we match the performance of an end-to-end model with ten times as many parameters. We consider 4 tasks and 4 datasets in Dutch and English.
\end{abstract}
\section{Introduction}
\label{sec:intro}
Spoken language understanding (SLU) generally follows a pipeline approach, where an automatic speech recognition (ASR) system is followed by a natural language understanding (NLU) module. These systems know multiple limitations, including cascading errors from one module to the next or discarding relevant information related to speech not captured by text. This motivated research directed towards end-to-end SLU that removed the explicit speech transcription required by older systems \cite{Serdyuk2018,Fluent}. End-to-end approaches concerned only with SLU tasks, however, struggle to match the performance of pipeline systems. Indeed, these specific datasets are considerably smaller and neural networks notably require large amounts of data to perform well. To circumvent this data scarcity issue, pretraining and multitask learning offer an elegant solution. Pretraining allows applying what a neural network learned from a large dataset to another domain by partly or entirely initializing the weights in the model of interest with the weights from the pretrained model. If the tasks share certain characteristics with the ultimate objective and the domain shift between the datasets is not too big, we can decrease considerably the amount of training examples and computation resources necessary to train the final model. During the actual training or finetuning of the model, a trade-off appears, as further training will inevitably lead to a specialization in favor of the new task and catastrophic forgetting of what was learned during the pretraining phase.
In contrast, with multitask learning, the tasks are trained in parallel. In this case, a similar trade-off occurs, caused by competing optimization objectives. This approach relies strongly on the assumption that the multiple objectives share a common ground and that the ability to solve one helps on solving others. It is easy to imagine many examples for which this assumption might hold or not, depending on the tasks and datasets considered. In this work, we investigate synergies between spoken language understanding and speech recognition. One task of interest is intent classification (IC), where a system must identify an intent with any number of arguments. A second task investigated here is sentiment classification, where the goal is to determine whether a spoken utterance expresses a positive, negative or neutral sentiment. Having a system able to transcribe speech to text is likely to be helpful for such tasks and requires fewer data and fewer parameters to reach a good performance \cite{Haghani2018,Rao2020}. 
These are two aspects that we will focus on in this work: We compare a multitask model optimized for both speech transcription and intent recognition or sentiment classification with a model pretrained on ASR then finetuned on SLU. We also look into edge cases in IC by simulating training sets with few examples. Further, we limit our model sizes considerably for the advantages in terms of speed and energy consumption that it brings.
Experiments are performed in English and in Dutch.

\section{Architecture}
\label{sec:model}
% \begin{figure}[ht]\centering
% \includegraphics[width=0.7\linewidth]{figures/architecture.png}
% \caption{Speech features are processed by the encoder. The ASR module is a hybrid CTC/attention transformer \cite{espnet}. The SLU module is a transformer with class attention (see \ref{sec:model:ic}). The dashed arrows correspond to different settings for the model.}
% \label{fig:architecture}
% \end{figure}
The model is composed of a generic transformer encoder on which we stack different modules, each specialized for one task. We experiment with different arrangements in order to find the optimal architecture that allows each task to benefit from - rather than compete with - each other. We have selected four tasks of interest: automatic speech recognition (ASR), intent classification (IC) and sentiment classification (SC). In this work, we explore the synergies between each task and ASR and leave other combinations for future works. Additionally, we explicitly want our models to be relatively small. Smaller models require less energy both during training and inference, which is good from a cost and environmental perspective.

\subsection{Encoder}
\label{sec:model:encoder}
The encoder is the only component that is entirely shared with all the tasks. It is composed of a VGG-like feature extractor followed by transformer encoder layers. The role of the convolutional frontend is two-fold: decrease the dimension along the frequency and time axes and learn the local relationships in the sequence. This has been successfully used in previous works to replace the positional encoding traditionally required by transformers \cite{Mohamed2020}. A sequence of speech features of length $T$ is reduced to a length $T' < T$ by the CNN, where $T' \approx T / 4$. The resulting features are further processed by a series of 12 self-attention and feedforward layers. The output of the encoder is a sequence of features of 256 units. It is used as input to the other modules and is never optimized in a standalone fashion. The encoder has under 18 million parameters. Wav2Vec 2.0 \cite{wav2vec2} has more than 300 million.

\subsection{Automatic Speech Recognition}
\label{sec:model:asr}
For the ASR module, we use a hybrid setting with a transformer decoder coupled with a CTC classification layer \cite{espnet}. The prediction units are 5,000 subword units. The CTC layer assigns to each acoustic unit a symbol that is either a token from the vocabulary or the blank symbol. A sentence can be derived either using a greedy algorithm that first removes duplicated symbols then blank tokens, or with a more fine-grained approach such as beam search. The transformer decoder generates one token at a time by comparing previously generated tokens with the acoustic evidence brought by the encoder. Following \cite{espnet}, the loss corresponding to this task combines a CTC loss \cite{Graves2006} and a cross-entropy loss. We also use label smoothing to avoid overconfident predictions. The total number of parameters for the ASR head is 13.3 million.

\subsection{Intent Classification}
\label{sec:model:ic}
Intent is composed of an action (e.g. move) and a number of arguments (e.g. where: forward, how: slowly). We encode the intent as a multi-hot vector where the first $k$ bits correspond to the $k$ actions, the following $l$ bits to the $l$ possible values for the first argument and so on. Inspired by \cite{ClassAtt}, we use multihead class attention to summarize the sequence of embeddings into one prediction. We introduce a special learned token, $x_\text{CLS}$, that is attending to each element in the sequence. In the original notation from \cite{Vaswani2017}, we substitute the embedded queries, $Q$, with $x_\text{CLS}$. However, contrarily to \cite{ClassAtt}, we exclude the CLS token from the keys and values. Class attention compares the class embeddings, $x_\text{CLS}$, to the sequence of representations, $X$, to identify and extract relevant information from the embedded sequence. The task loss is binary cross-entropy. The intent classification module has only 825 thousand parameters.

\subsection{Sentiment Classification}
\label{sec:model:sc}
For sentiment classification, we use the same architecture as in Section \ref{sec:model:ic}. The loss for this task is cross-entropy. We apply label smoothing to avoid overconfident predictions.

% \subsection{Named Entity Recognition}
% \label{sec:model:ner}
% Instead of predicting a sequence of entities as in \cite{SLUE}, we choose to adopt another approach, where we try to assign a symbol to each of the output units of the ASR module. We follow the BIO framework, where a symbol is composed of 
% the NER module is composed of transformer encoder layers followed by a classification layer to classify the token as being a part of an entity or not. We follow the BIO-scheme which assigns a label based on whether a token marks the beginning of an entity (B), is part of one (I) or outside (O). The input to the NER modules are hidden representations from the ASR decoder. Consequently, one element in the input sequence corresponds to one wordpiece unit.

\section{Methodology}
\label{sec:methods}
In this section, we describe how we train and evaluate the models. For each language, we have selected one ASR dataset to pretrain the encoder and ASR decoder and two datasets to perform multitask learning (MTL). Pretraining has shown its benefits in numerous applications and serves here as a shortcut to save considerable computing resources by initializing parts of our model. We pretrain our model with ESPnet's recipe for ASR. At this stage, we freeze the encoder’s parameters, as our experiments did not show improvements when the encoder was updated. The decoder is trained with teacher forcing, meaning that the true tokens are used as input rather than the decoder's own prediction. Although this introduces a difference between training and inference, it also makes the training procedure faster by allowing parallel training, and more stable by avoiding introducing errors while the encoder is still learning.

\subsection{Training}
\label{sec:methods:training}
All downstream datasets have at least two targets: transcriptions for ASR and labels for IC and SC. Given a speech utterance, the model predicts both targets. Then, the weighted sum of the errors relative to each task is backpropagated through the network and the parameters are updated accordingly. The decoder is thus trained jointly on ASR and either IC or SC. We set the loss weights experimentally by measuring the performance on the validation set.

\subsection{Evaluation}
\label{sec:methods:evaluation}
We evaluate the models by measuring the task performance on a left-out subset. For intent classification, we report the accuracy. Note that for calculating the accuracy, we consider that a prediction is correct if all the elements are correct -- intent and arguments. For sentiment classification, we report the macro averaged f1-score.\\
In addition, we explore the ability of our model to learn in data shortage scenarios, using as few training examples as possible. We follow \cite{Renkens2014}'s direction and train our multitask models with varying amount of training data for one speaker. In practice, for each speaker dataset, we define several training sets of increasing sizes to train and evaluate the models. This allows us to build learning curves that show the average performance of the same model on increasing training sizes. We compare these results with two types of baseline features: one is obtained by encoding the gold transcription with a pretrained BERT (NLP in Figure \ref{fig:results}), and the second is obtained by transcribing speech utterances with an ASR model and encoding them with a pretrained BERT (Pipeline in Figure \ref{fig:results}).

\section{Experiments}
\label{sec:exp}
We explore datasets in both English and Dutch. We use the \textit{Corpus Gesproken Nederlands (CGN)} \cite{CGN} and \textit{Librispeech} \cite{Librispeech} for pretraining. The 80-dimensional MEL-filterbanks are pre-computed for all the datasets. We modified \textit{ESPnet} \cite{espnet} to fit our specific requirements. For the actual training of the multitask models, we use 2 datasets in Dutch and 2 datasets in English. 

\subsection{Datasets}
\label{sec:exp:data}
\paragraph{Corpus Gesproken Nederlands} \cite{CGN} is a collection of recordings in Dutch and Flemish collected from various sources totaling more than 900 hours. They are divided into 15 components (`a' to `o') based on their nature. After removing short and overlapping utterances, we divide each component into three subsets that will serve as training, validation and test sets. We leave out three components (`a', `c' and `d') because they differ considerably from the other ones. The subsets from the remaining components, totaling 415 hours of speech, are concatenated together to form the training, validation and test sets.\\
\paragraph{Librispeech} \cite{Librispeech} contains about a thousand hours of read English speech derived from audiobooks. The authors provide official splits for training, validating and testing the models and this dataset serves as reference in ASR. We use all 960 hours for training.\\
\paragraph{Grabo} \cite{Renkens2014} is composed of spoken commands, mostly in Dutch (one speaker speaks in English) intended to control a robot. There are 36 commands that were repeated 15 times by each of the 11 speakers. More precisely, the robot can perform eight actions, each being further defined by some attributes (e.g., the robot can move forward or backward, rapidly or slowly). We follow the same methodology as \cite{Renkens2018}.\\
\paragraph{Patience Corpus} \cite{Patcor} is derived from a card game where the player is giving vocal instructions to move cards between different stacks on the table. The dataset contains recordings from 8 different Flemish speakers. The intent of the player is represented as a multi-hot vector. Again, we use the same splitting methodology as \cite{Renkens2018}.\\
\paragraph{Fluent Speech Commands} \cite{Fluent} contains 30,043 utterances from 97 English speakers. Each utterance is a command to control smart-home appliances or virtual assistant. We use the challenge splits proposed in \cite{MASE}. They propose two test sets, one with unknown speakers (FSC-S) and one with unknown utterances (FSC-U)\\
% \paragraph{SLUE-VoxPopuli} was annotated and proposed as a benchmark in \cite{SLUE} for the named entity recognition task. The test set is not released, hence we evaluate on the validation set.\\
\paragraph{SLUE-VoxCeleb} \cite{SLUE} is a benchmark dataset for sentiment classification. The task is to classify each utterance as being positive, negative or neutral.

\subsection{Model Selection}
\label{sec:exp:ablation}
\begin{figure}[ht]
    \centering
    \begin{subfigure}[b]{0.49\linewidth}
        \centering
        \includegraphics[width=\linewidth,height=3cm]{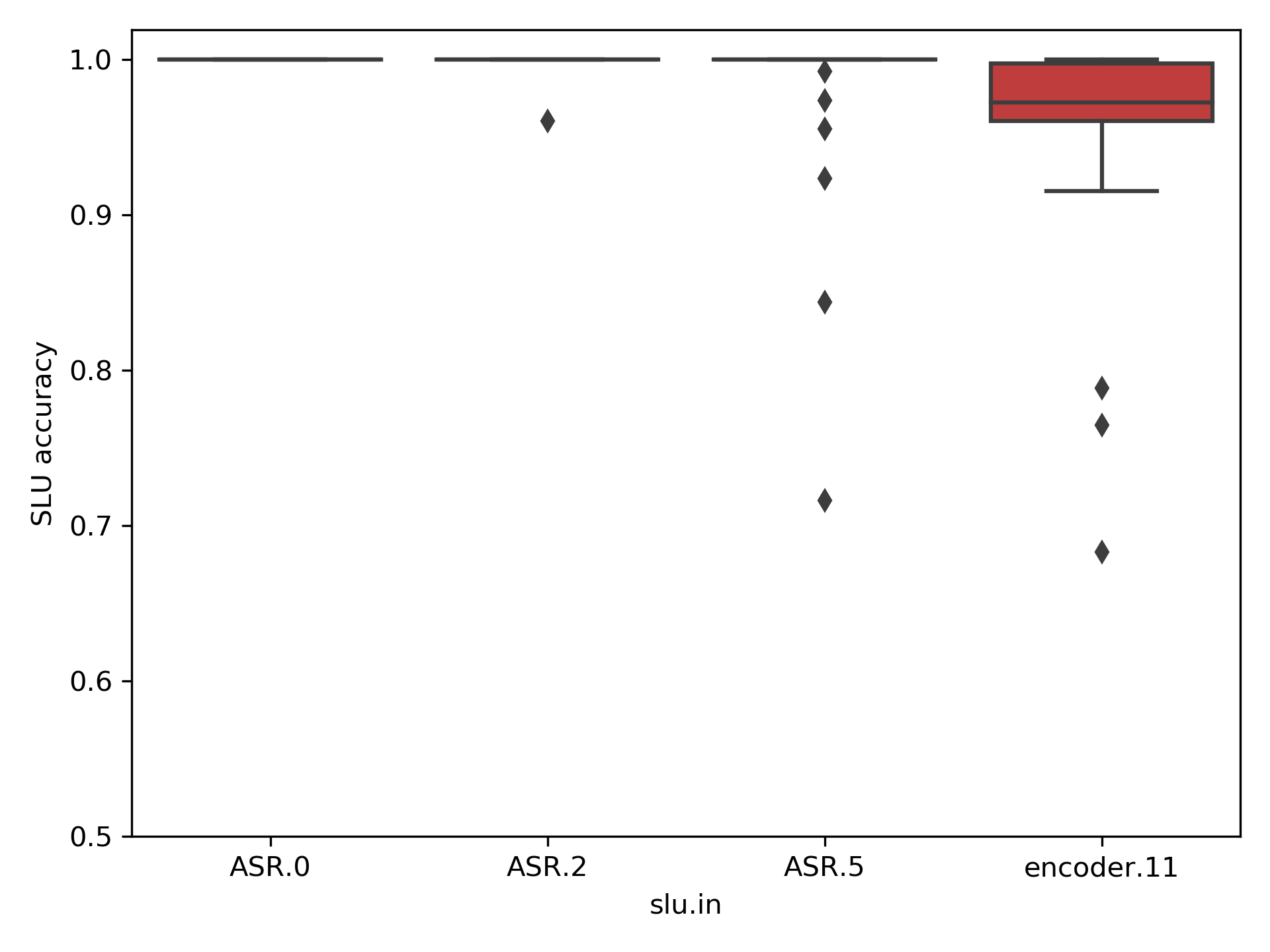}
        \caption{\textit{Multitask Learning}}
        \label{fig:ablation:grabo-mtl}
    \end{subfigure}
    \begin{subfigure}[b]{0.49\linewidth}
        \centering
        \includegraphics[width=\linewidth,height=3cm]{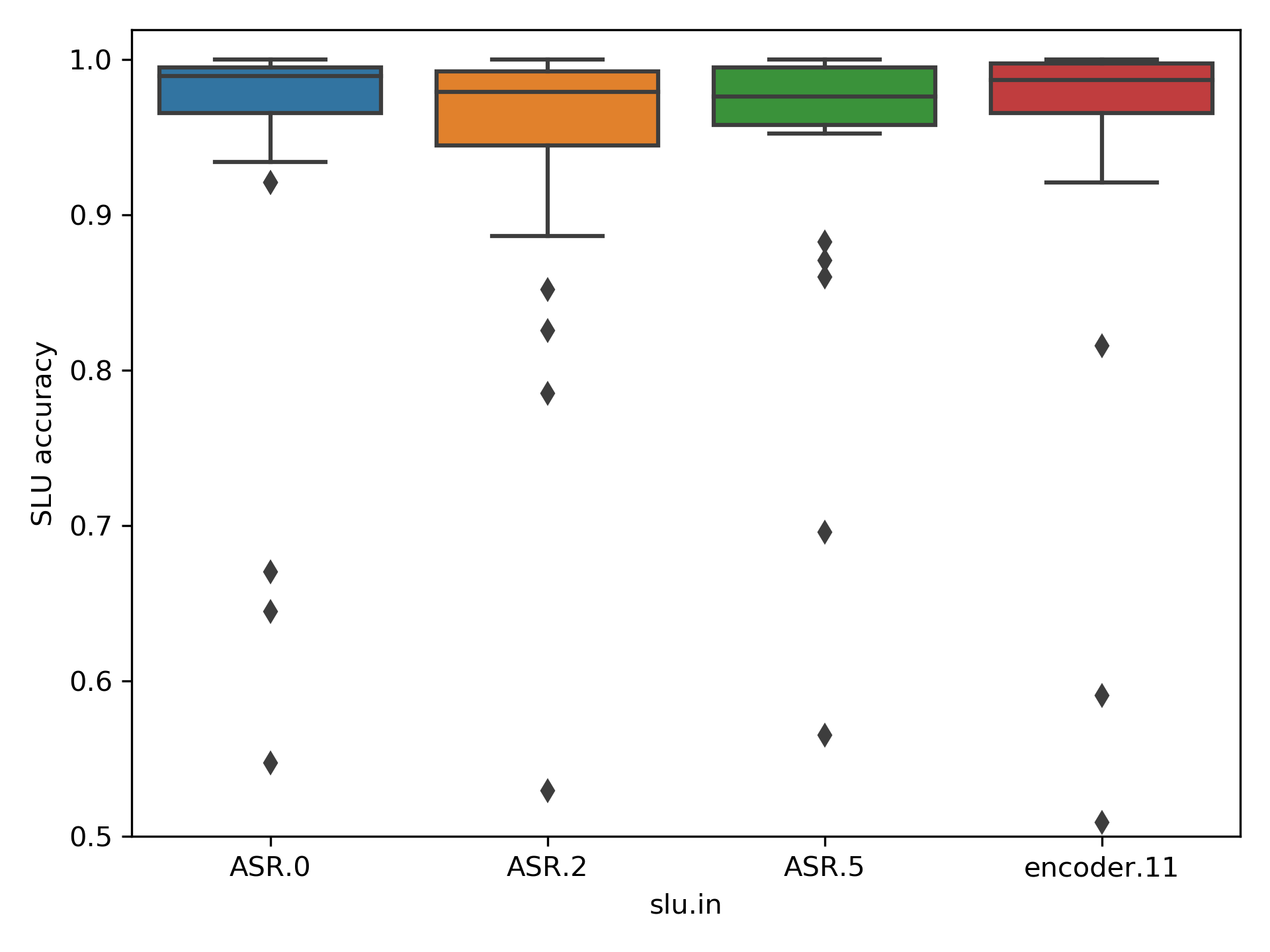}
        \caption{\textit{Single task (SLU)}}
        \label{fig:ablation:grabo-slu}
    \end{subfigure}
    \caption{Comparison of different settings on Grabo for the multitask model where the SLU module is connected to the output of the encoder (encoder.11) or the first (ASR.0), third (ASR.2) or last layer (ASR.5) of the ASR decoder. Each box represents the distribution of the score across speakers and folds where each model was trained on a dataset with 2 examples per class. Performance of a multitask model trained on ASR+SLU (left) and a model trained solely on SLU (right) is shown.}
    \label{fig:ablation}
\end{figure}

In this experiment, we compare the model performance on the SLU task depending on the features it receives, namely intermediary representations from the first, third or last layer of the ASR decoder or from the encoder's output. We perform the experiment first on Grabo dataset where we build small training sets with 2 examples per class to simulate a data shortage scenario. All the models are initialized with the same pretrained weights to ensure comparable results.
We perform this operation three times and report the accuracy in Figure \ref{fig:ablation}. The model where the SLU module connects to the encoder's output shows the largest variability and the lowest average performance. The encoder's output corresponds to acoustic units that do not have much to do with language. The decoder, however, produces a sequence of subword units given the acoustic evidence and the previous predicted tokens. When the SLU module is connected to the encoder's output, it must summarize a longer sequence where each element conveys acoustic rather than linguistic information. In contrast, aside from the obvious advantage that the SLU head must cope with shorter sentences, the implicit language model in the decoder gives another meaning to the tokens that is dictated by the masked language modelling objective. These representations consider the acoustics through the cross-attention but also linguistic elements with the self-attention. 
The model trained on features from the last layer of the decoder shows an increased number of outliers, depicted by the diamonds in Figure \ref{fig:ablation}. This indicates that there might be a trade-off between using better-defined or lower-level features as we change the capacity of the ASR module dedicated to only one task. 
In other words, when the SLU head is connected to the last decoder layer, the decoder has fewer parameters to learn elements necessary to ASR but detrimental, or unnecessary to SLU. Consequently, competition arises between both objectives.\\
On Figure \ref{fig:ablation}, the performance of the models trained with a unique objective is lower than when we optimize for both ASR and SLU. This confirms that performing both tasks helps the performance on SLU. The difference between the models' performance in Figure \ref{fig:ablation:grabo-slu} is not statistically significant, even when the encoder's output is used. This indicates that the improved performance between the models is due to the dual objective rather than the increased capacity given by the additional layers.

\subsection{Data Shortage Scenario}
\label{sec:exp:lowresource}
\begin{figure}
    \centering
    \begin{subfigure}[b]{0.49\linewidth}
        \centering
        \includegraphics[width=\linewidth,height=3cm]{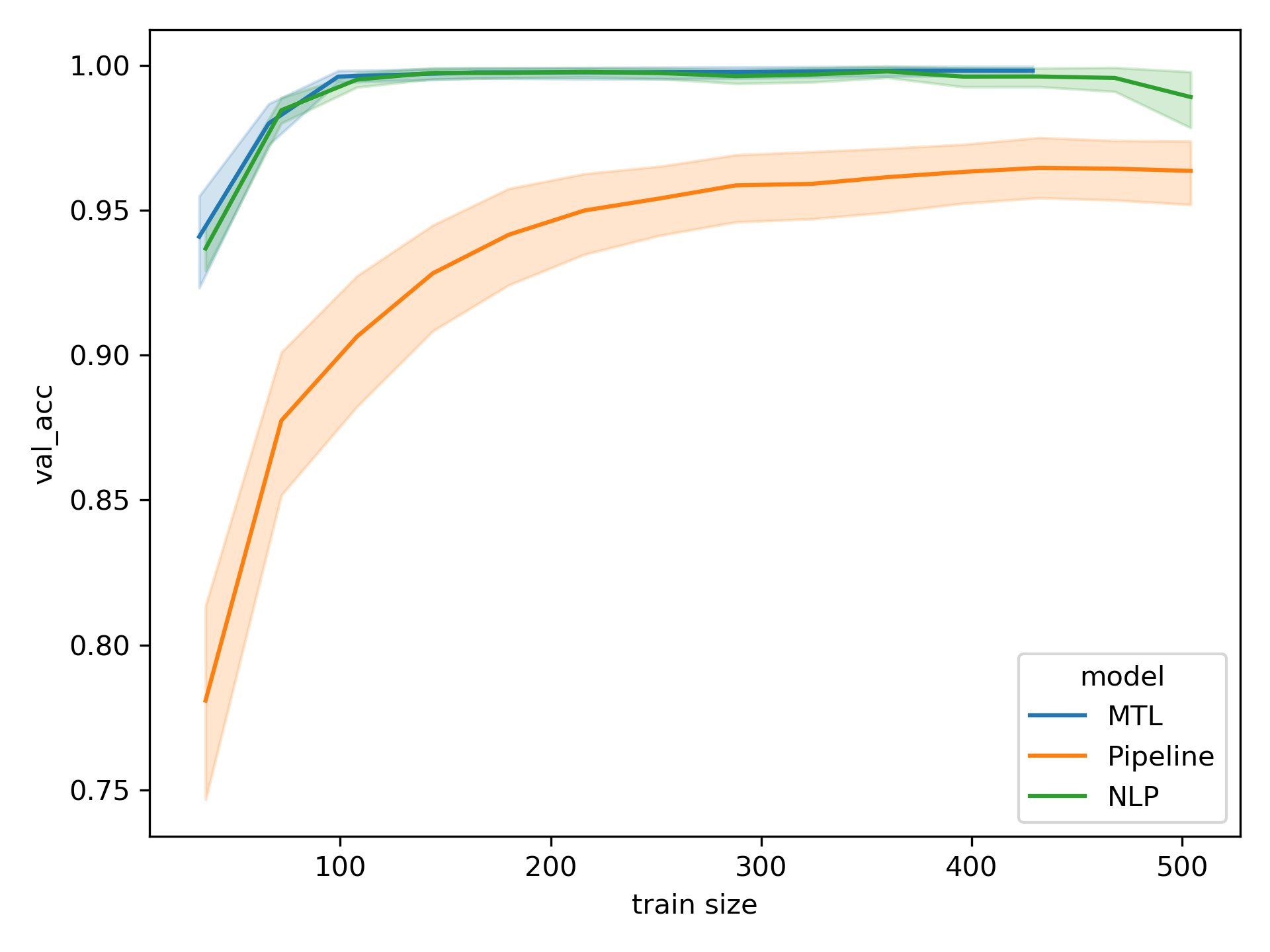}
        \caption{\textit{Grabo}}
        \label{fig:results:grabo}
    \end{subfigure}
    \begin{subfigure}[b]{0.49\linewidth}
        \centering
        \includegraphics[width=\linewidth,height=3cm]{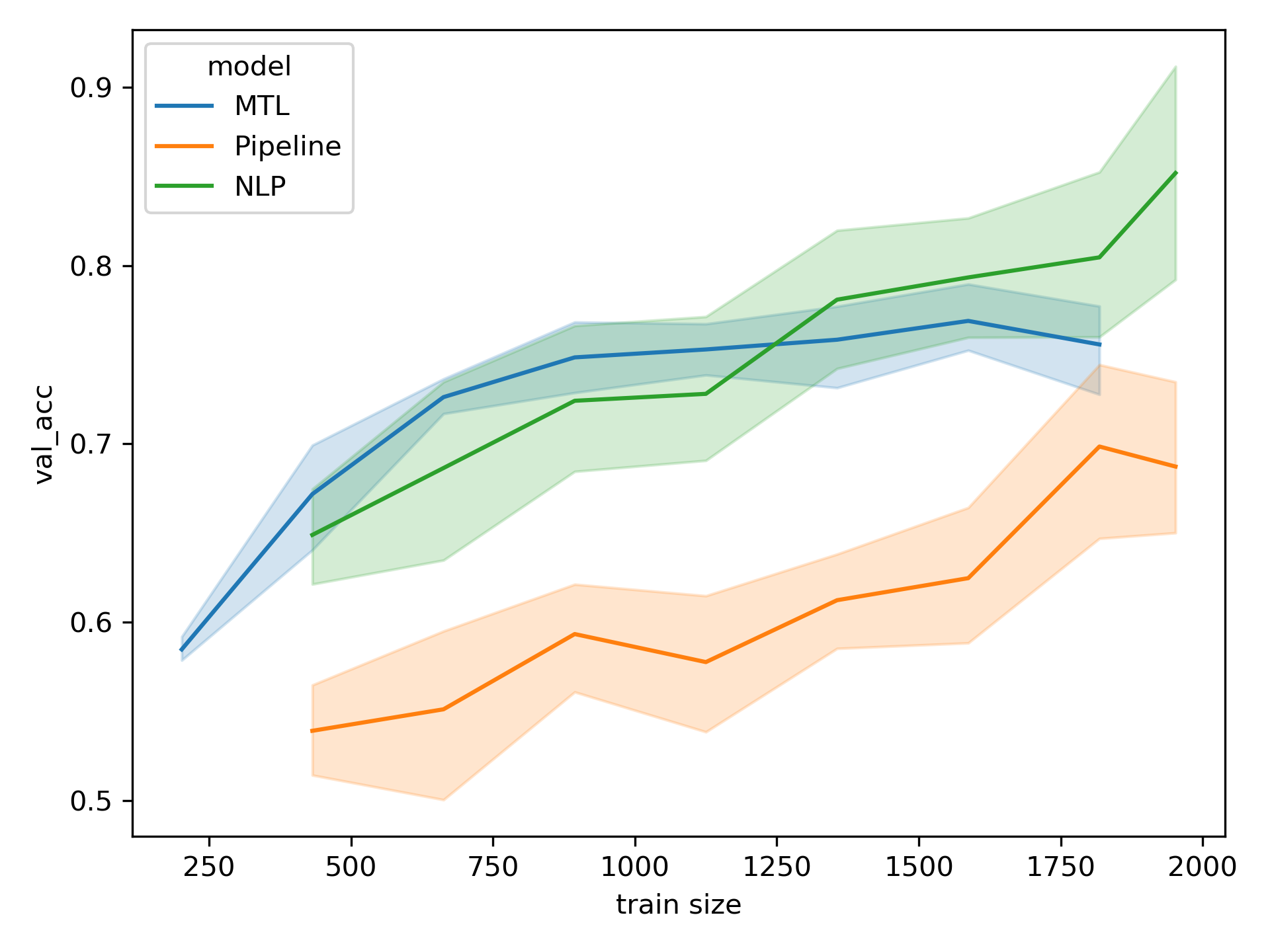}
        \caption{\textit{Patience Corpus}}
        \label{fig:results:patcor}
    \end{subfigure}
    \caption{
    Comparison of different models trained with increasing train set sizes on \textit{Grabo} (left) and \textit{Patcor} (right). Each curve represents the accuracy on the test set as a function of the size of the training set. The bands around the curves represent a 68\% confidence interval.}
    \label{fig:results}
\end{figure}
When data is scarce, multitask learning can do a lot, as showed in Figure \ref{fig:results}. We compare our multitask model (blue) with two baselines: we either use ASR transcripts or gold transcription to generate features with BERTje \cite{wietsedv2019}. These sequences of representations are used as input to train a SLU module. As the Grabo dataset contains the exact same set of sentences repeated multiple times, we do not finetune BERT in order not to make the task trivial. We observe remarkable results on both datasets, Grabo and Patience, especially in low resource scenario where our model shows impressive performance compared to the baselines. The Patience Corpus is considerably more complicated than Grabo because of the diversity of training examples and the larger number of possible outputs. Nonetheless, our model shows a better performance for small training sizes although for larger training sets, the NLP model's accuracy score rises above the results of the multitask model.

\subsection{Results}
\label{sec:exp:results}
% \begin{table}[ht]
%     \small
%     \centering
%     \begin{tabular}{l|c|c|c|c|c}
%           & Grabo & FSC-S & FSC-U & VP & VC \\
%          Model & acc & acc  & acc & f1 & f1 \\
%          \hline
%          MTL & 100 & 98.2 & 88.1 &  & 75.75 (dev) \\
%          ESPnet-SLU \cite{espnet-slu} & 97.2 & 97.5 & 78.5 & -- & -- \\
%          NLP \cite{SLUE} & -- & -- & -- & 81.4 & 64.3 \\
%          Pipeline \cite{SLUE} & -- & -- & -- & 57.8 & 63.3 \\
%          End-to-End \cite{SLUE} & -- & -- & -- & 50.9 & 48.5 \\
%     \end{tabular}
%     \caption{Performance on the test set of our model compared to baselines. FSC: Fluent Speech Commands, VP: VoxPopuli, VC: VoxCeleb. NLP, Pipeline and End-to-End refer to the provided baselines with no language model \cite{SLUE}.}
%     \label{tab:results}
% \end{table}
\begin{table}[ht]
    \small
    \centering
    \begin{tabular}{l|c|c|c}
         Model & Grabo & FSC-S & FSC-U \\
         \hline
         MTL & 100 & 98.2 & 88.1 \\
         ESPnet-SLU \cite{espnet-slu} & 97.2 & 97.5 & 78.5 \\
    \end{tabular}
    \caption{Accuracy of the models on the test set of our model compared to the baseline proposed in \cite{espnet-slu}. FSC-S and FSC-U correspond to the Fluent Challenge splits, respectively with unknown speakers and with unknown utterances.}
    \label{tab:results:ic}
\end{table}
\begin{table}[ht]
    \small
    \centering
    \begin{tabular}{l|c|c|c|c|c}
         Model & \# params & macro-f1 \\
         \hline
         MTL (Ours) & 31  M & 49.82 \\
         NLP & 390 M & 64.3 \\
         Pipeline & 707 M & 63.3 \\
         E2E & 317 M & 48.5 \\
    \end{tabular}
    \caption{Performance on the test set of our model on sentiment classification (VoxCeleb) compared to the baseline \cite{SLUE}.}
    \label{tab:results:slu}
\end{table}
Table \ref{tab:results:ic} compares our results with the baseline models proposed in \cite{espnet-slu} on Fluent and Grabo. As our models are pretrained on a large variety of speakers, it is not surprising that they are robust to unseen speakers. However, it seems that the combination of pretraining and multitask learning also considerably helps the model to generalize to unseen output patterns, as evidenced by the strong score on FSC-U. At this stage, we did not observe strong evidence that SLU positively impacts ASR scores. In Table \ref{tab:results:slu}, we compare the performance of our model on the sentiment classification task. The model labelled \textit{NLP} corresponds to the performance of DeBERTa large \cite{deberta} on the gold transcriptions. \textit{Pipeline} is composed of Wav2Vec 2.0 large \cite{wav2vec2} followed by DeBERTa large. E2E correspond to Wav2Vec 2.0 without a dedicated language model for decoding. The results shown here are reported in \cite{SLUE}. Although our model only slightly improves over the end-to-end baseline, it does so with considerably fewer parameters.

\section{Conclusion}
\label{sec:conclusion}
In this work, we proposed a multitask-learning model to do both automatic speech recognition and either intent classification or sentiment classification in parallel. We see an improvement over doing IC independently, even more so in low-resource scenario where it is beneficial to access knowledge from written as well as spoken language. Further, we reach a better performance than the end-to-end baseline on sentiment classification, although we are using a model that is ten times smaller. In the future, we will keep working on this topic by including more tasks to train in parallel. Additionally, we did not mention the impact of MTL on ASR results, which we will explore in more details. Finally, we used in this work class attention, defined in Section \ref{sec:model:ic}. It has proved to be a very efficient addition to our model but requires an article of its own to explore all its benefits.

\clearpage
\bibliographystyle{IEEEtran}
\bibliography{strings,refs}

\end{document}